\documentclass[letterpaper, 10 pt, journal, twoside]{ieeetran}

\usepackage{multirow}
\usepackage{graphicx}
\usepackage{pifont}
\usepackage{amsmath,amssymb,amsfonts}
\usepackage{arydshln}
\usepackage{longtable}
\usepackage{booktabs}
\usepackage{textcomp}
\usepackage{subfigure}
\usepackage{stfloats}
\usepackage{url}
\usepackage{verbatim}
\usepackage{cite}
\usepackage{textcomp}
\usepackage[colorlinks=true,linkcolor=black,citecolor=blue,urlcolor=blue,]{hyperref}

\graphicspath{{Figure/}}
\newcommand{\etal}{et~al.\ }

\begin{document}

\setlength{\abovedisplayskip}{4pt}  % default value is 12pt
\setlength{\belowdisplayskip}{4pt}

\title{EF-Calib: Spatiotemporal Calibration of \\ Event- and Frame-Based Cameras \\ Using Continuous-Time Trajectories}

\author{Shaoan~Wang,
        Zhanhua~Xin,
        Yaoqing~Hu,
        Dongyue~Li,
        Mingzhu~Zhu,
       and Junzhi~Yu,~\IEEEmembership{Fellow,~IEEE}
           \thanks{This work was supported in part by the National Natural Science Foundation of China under Grant T2121002 and Grant 62233001, and in part by the Beijing Natural Science Foundation under Grant 2022MQ05. \emph{(Corresponding author: Junzhi Yu.)}}
        \thanks{Shaoan~Wang, Zhanhua~Xin,  Yaoqing~Hu, Dongyue~Li, and Junzhi~Yu are with the State Key Laboratory for Turbulence and Complex Systems, Department of Advanced Manufacturing and Robotics, College of Engineering, Peking University, Beijing 100871, China (e-mail: wangshaoan@stu.pku.edu.cn; xinzhanhua@stu.pku.edu.cn; 2101111894@stu.pku.edu.cn; 2001111648@stu.pku.edu.cn; junzhi.yu@ia.ac.cn).}
        \thanks{Mingzhu~Zhu is with the Department of Mechanical Engineering, Fuzhou University, Fuzhou 350000, China (e-mail: mzz@fzu.edu.cn).}
}

\markboth{IEEE Robotics and Automation Letters. Preprint Version.}
{Wang \MakeLowercase{\textit{et al.}}: EF-Calib: Spatiotemporal Calibration of Event- and Frame-Based Cameras Using Continuous-Time Trajectories}

\maketitle

\begin{abstract}
The event camera, a bio-inspired asynchronous triggered camera, offers promising prospects for fusion with frame-based cameras owing to its low latency and high dynamic range. However, calibrating stereo vision systems that incorporate both event- and frame-based cameras remains a significant challenge. In this letter, we present EF-Calib, a spatiotemporal calibration framework for event- and frame-based cameras using continuous-time trajectories. A novel calibration pattern applicable to both camera types and the corresponding event recognition algorithm are proposed. Leveraging the asynchronous nature of events, a derivable piece-wise B-spline to represent camera pose continuously is introduced, enabling calibration for intrinsic parameters, extrinsic parameters, and time offset, with analytical Jacobians provided. Various experiments are carried out to evaluate the calibration performance of \mbox{EF-Calib}, including calibration experiments for intrinsic parameters, extrinsic parameters, and time offset. Experimental results demonstrate that EF-Calib outperforms current methods by achieving the most accurate intrinsic parameters, comparable accuracy in extrinsic parameters to frame-based method, and precise time offset estimation. EF-Calib provides a convenient and accurate toolbox for calibrating the system that fuses events and frames. The code of this paper is open-sourced at: \href{https://github.com/wsakobe/EF-Calib}{https://github.com/wsakobe/EF-Calib}.
\end{abstract}


\IEEEpeerreviewmaketitle

\section{Introduction}
\IEEEPARstart{I}{n} recent years, there has been a growing interest among researchers in a novel bio-inspired camera called the event camera \cite{Bib:Gallego2020TPAMI}. Abandoning the frame-triggered concept of conventional cameras, each pixel of the event camera can be considered as responding independently and asynchronously to changes in illumination, resulting in an extremely low-latency and high-dynamic-range response pattern. These advantages offer competitive prospects for event cameras in areas such as robotics \cite{Bib:Zhu2023IROS}, autonomous driving \cite{Bib:Zhou2023ICRA, Bib:Li2024ICRA}, VR/AR \cite{Bib:Jiang2024TPAMI}, and camera imaging \cite{Bib:Han2024TPAMI}.

\begin{figure}[!t]
\centering
\includegraphics[width=.46\textwidth]{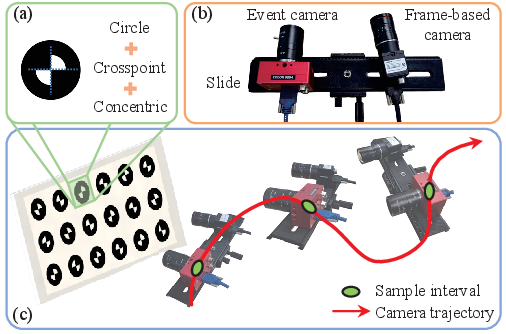}
\caption{Overview diagram of EF-Calib. (a) The novel calibration pattern consists of the concentric circle and crosspoint. (b) The stereo vision system consists of an event camera and a frame-based camera. (c) The calibration process of EF-Calib.}
\label{Fig:real-world}
\end{figure}

However, due to the imaging principle of event cameras, they only respond to changes in illumination, making it challenging to capture absolute grayscale or RGB values like frame-based cameras. This limitation weakens the ability of event cameras to perceive and understand the environment. Therefore, many recent studies have attempted to fuse events with images in order to fully utilize the unique advantages of both modalities, as illustrated in Fig.~\ref{Fig:real-world}. Some novel SLAM systems achieve more robust localization under fast motion by fusing events and frames \cite{Bib:Zhu2023IROS, Bib:Guan2023TASE, Bib:Lee2024RAL, Bib:Vidal2018RAL}. In addition, several studies are exploring how to fuse events and frames for object detection in challenging environments \cite{Bib:Zhou2023ICRA, Bib:Luo2024RAL}. In recent years, some event-centric datasets with multiple sensors, including frame-based cameras, have also been widely proposed \cite{Bib:Gao2022RAL, Bib:Creb2024TIV, Bib:Chen2024TIV}.

It is important to note that calibrating the intrinsic and extrinsic parameters of each camera is an indispensable step in the context of multi-camera fusion. Classical multi-camera calibration frameworks typically require time synchronization among all cameras, followed by the acquisition of each camera’s parameters through the synchronous capture of images of a calibration board in various poses \cite{Bib:Zhang2000TPAMI}. However, due to the asynchronous nature of event cameras, it is challenging to aggregate multiple events into single ``frames'' and time-synchronize them with traditional images. Additionally, events are generated only when the illumination at each pixel changes. As a result, a static event camera cannot capture static objects without any illumination change. In conclusion, a user-friendly, and cost-effective calibration framework is needed for systems that include both event- and frame-based cameras.

To address the aforementioned issues, this letter proposes a novel spatiotemporal calibration framework for event- and frame-based cameras called EF-Calib. To the best of our knowledge, EF-Calib is the first calibration framework to achieve joint calibration of event- and frame-based cameras without requiring any time synchronization. The main contributions of this paper are as follows:

\begin{enumerate}
    \item A novel spatiotemporal calibration framework for event- and frame-based cameras is proposed. This framework can obtain the intrinsic and extrinsic parameters, as well as the time offset without requiring any hardware synchronization.

    \item Leveraging the asynchronous and low-latency properties of the event camera, the framework introduces a continuous-time trajectory to optimize its motion trajectory, facilitating arbitrary timestamp alignment with the frame-based camera.

    \item Extensive experiments are conducted in diverse scenarios to validate the proposed calibration framework. The results indicate that the framework achieves accuracy close to that of frame-based camera calibration methods and consistently calibrates the time offset between the cameras.
\end{enumerate}

The rest of the letter is organized as follows. Sec.~II summarizes the related works. Sec.~III presents the preliminaries of event-based vision and continuous-time trajectory. Sec.~IV introduces the calibration framework. Sec.~V evaluates the calibration performance from different aspects. At last, Sec.~VI presents the conclusion of this letter.

\section{Related Works}
For geometric vision, camera calibration is particularly crucial as it serves as the initial step in processing the input image signal, with the quality of calibration often dictating the performance of subsequent tasks. Traditional camera calibration has undergone significant evolution. The most prevalent calibration method today involves capturing images of a calibration pattern with a known size, such as a checkerboard, from various viewpoints to identify corresponding feature points \cite{Bib:Zhang2000TPAMI}. Subsequently, the intrinsic and distortion parameters of each camera, as well as the extrinsic parameters between cameras, are automatically calculated.

Nevertheless, applying this static and discrete calibration method to event cameras, which are triggered by changes in illumination or relative motion, presents challenges. Initially, many open-source event camera calibration kits utilized a synchronized blinking LED calibration board \cite{Bib:RPGCalib, Bib:OrchardCalib, Bib:VLOCalib}, enabling alignment between cumulative event camera frames and traditional images. However, this approach requires complex hardware, leading to higher costs and reduced calibration accuracy. Alternatively, static calibration can be achieved by adjusting external light intensity, but this introduces significant noise and fails to account for time offset between cameras. As a result, these toolkits are not the most efficient solution for calibrating event-based and frame-based cameras.

In recent years, there has been increased focus on designing new calibration frameworks to facilitate event camera calibration using existing calibration boards. Muglikar \etal \cite{Bib:Muglikar2021CVPR} utilize deep learning-based image reconstruction networks, such as E2VID \cite{Bib:Rebecq2021TPAMI}, to record events generated by moving the calibration board and then apply the reconstructed images to classical calibration methods. However, these methods heavily rely on the quality of image reconstruction and face challenges in achieving time synchronization with conventional cameras. Another approach involves directly utilizing events generated during camera motion for camera calibration. Huang \etal \cite{Bib:Huang2021IROS} proposed a calibration framework based on a circular calibration board and employed B-splines to optimize the movement trajectory, which is the most similar method to the one proposed in this letter. However, they directly use clustered asynchronous events as features for optimization, compromising sub-pixel accuracy and being highly sensitive to noise. Salah \etal \cite{Bib:Salah2023Arxiv} also utilize circular calibration boards and introduce eRWLS to fit circular features with sub-pixel accuracy. However, they compress events over a period into a fixed timestamp to obtain a reference ``frame'' making this method challenging to synchronize with a frame-based camera. Furthermore, it does not account for the deformation of circular features at different viewing angles, leading to reduced sub-pixel localization accuracy. The calibration framework proposed in this letter continues this concept and provides an improvement to address the problems of these methods.

\section{Preliminaries}
\subsection{Event-Based Vision}
Unlike conventional cameras, each pixel of the event camera is independently triggered and responds to changes in the logarithmic illumination signal $L(\mathbf{u}_k,t_k)$. An event $(\mathbf{u}_k,t_k, p_k)$ is triggered when the change in logarithmic illumination received by a pixel $\mathbf{u}_k=(x_k, y_k)$ exceeds a threshold value $C$, i.e.,
\begin{equation}
    \label{Eqn:1}
    \Delta L(\mathbf{u}_k,t_k)\doteq L(\mathbf{u}_k,t_k)-L(\mathbf{u}_k,t_k-\Delta t)=p_kC
\end{equation}
where $\Delta t$ is the time since the last triggered event by the same pixel, $p_k \in \{-\text{1}, +\text{1}\}$ is the polarity of the event.

\subsection{Continuous-Time Trajectory Representation}
Continuous-time trajectories are often represented utilizing a weighted combination of the temporal basis functions \cite{Bib:Rehder2016TRO}, such as polynomial functions, FFTs, and Bézier curves. In this letter, the uniform B-spline is introduced as a representation of the continuous-time trajectory. B-splines have the advantages of smoothness, local support, and analytic derivatives, which are well-suited for representing the 6-DoF pose of the event camera \cite{Bib:Mueggler2018TRO, Bib:Huai2022SensorsJ, Bib:Patron2015IJCV}. Following the formulation of cumulative $k$th degree B-spline $\mathcal{L}$, the event camera pose $\mathbf{T}^e_w(\tau) \in\mathbb{SE}\text{3}$ at any time $\tau \in [t_i, t_{i+1})$, it can be represented by $N$ control points $\mathbf{T}_i \in\mathbb{SE}\text{3}, i \in[\text{0},\text{1},\ldots, N-\text{1}]$:
\begin{equation}
    \label{Eqn:2}
    \mathcal{L}: \mathbf{T}_w^e(\tau)=\mathbf{T}_i\cdot\prod_{j=\text{1}}^{k}\mathrm{Exp}\left(\tilde{\mathbf{B}}_{j}(\tau)\cdot\mathrm{Log}\left(\mathbf{T}_{i+j-1}^{-1}\mathbf{T}_{i+j}\right)\right)
\end{equation}
where $\tilde{\mathbf{B}}_{j}(\tau)$ is the cumulative basis function, which is denoted by
\begin{equation}
    \label{Eqn:3}
    \tilde{\mathbf{B}}_{j}(\tau)=\tilde{\mathbf{M}}^{(k)}\mathbf{u}
\end{equation}

\begin{equation}
    \label{Eqn:1}
    \mathbf{u}=\begin{bmatrix}1&u&\cdots&u^k\end{bmatrix}^T, u=\frac{\tau-t_i}{t_{i+1}-t_i}
\end{equation}
where $\tilde{\mathbf{M}}^{(k)} \in \mathbb{R}^{(k+\text{1})\times (k+\text{1})}$ is the cumulative blending matrix of B-splines. Since the control points of the B-splines are uniformly distributed on the time scale, the cumulative blending matrix $\mathbf{M}^{(k)}$ is constant. In this letter, considering the continuity and complexity, we use cubic B-splines to represent the camera pose, i.e., $k = \text{3}$. The corresponding cumulative mixing matrix $\tilde{\mathbf{M}}^{(3)}$ can be calculated as described in \cite{Bib:Qin2000VC}.

Representing B-splines using (\ref{Eqn:2}) leads to computational inefficiency and coupling issues. According to \cite{Bib:Sommer2020CVPR, Bib:Ovren2019IJRR}, decoupling $\mathbf{T}_w^e(\tau) \in \mathbb{SE}$3 into rotations $\mathbf{R}_w^e(\tau) \in \mathbb{SO}$3 and translations $\mathbf{t}_w^e(\tau) \in \mathbb{R}^\text{3}$ improves both motion trajectory representation and computational time. Hence, the continuous-time trajectory of the camera pose can be finally formulated as
\begin{equation}
    \label{Eqn:6}
    \mathbf{R}_w^e(\tau)=\mathbf{R}_i\cdot\prod_{j=\text{1}}^{3}\mathrm{Exp}\left(\tilde{\mathbf{B}}_{j}(\tau)\cdot\mathrm{Log}\left(\mathbf{R}_{i+j-1}^{-1}\mathbf{R}_{i+j}\right)\right)
\end{equation}

\begin{equation}
    \label{Eqn:7}
    \mathbf{t}_w^e(\tau)=\mathbf{t}_i+\sum_{j=\text{1}}^{3}\tilde{\mathbf{B}}_{j}(\tau)\cdot(\mathbf{t}_{i+j}-\mathbf{t}_{i+j-1})
\end{equation}

After decoupling the pose trajectory into two cubic B-splines, the corresponding analytic derivatives \cite{Bib:Sommer2020CVPR} can also be derived
\begin{equation}
\label{Eqn:8}
\begin{aligned}
    \dot{\mathbf{R}_w^e}(\tau)& =\mathbf{R}_w^e(\tau)\cdot\left(\boldsymbol{\omega}^{(\text{3})}(\tau)\right)_{\wedge}  \\
    &=\mathbf{R}_{i}\left(\dot{\mathbf{A}}_\text{1}\mathbf{A}_\text{2}\mathbf{A}_\text{3}+\mathbf{A}_\text{1}\dot{\mathbf{A}}_\text{2}\mathbf{A}_\text{3}+\mathbf{A}_\text{1}\mathbf{A}_\text{2}\dot{\mathbf{A}}_\text{3}\right)
\end{aligned}
\end{equation}

\begin{equation}
    \label{Eqn:9}
    \mathbf{v}_e^w(\tau)=\dot{\mathbf{t}}_e^w(\tau)=\mathbf{t}_i\cdot\sum_{j=\text{1}}^\text{3}\dot{\tilde{\mathbf{B}_{j}}}(\tau)\cdot(\mathbf{t}_{i+j}-\mathbf{t}_{i+j-1})
\end{equation}
where $\mathbf{A}$, $\dot{\mathbf{A}}$, and $\dot{\tilde{\mathbf{B}}}$ can also be found in \cite{Bib:Sommer2020CVPR}. Since this letter utilizes the uniform B-spline, $\Delta t$ is equal to the time interval between any two consecutive knots, i.e., $\Delta t = t_{i+1} - t_i$.

\begin{comment}
where
\begin{equation}
    \label{Eqn:10}
    \mathbf{A}_{j}=\mathrm{Exp}\left(\tilde{\mathbf{B}}_{j}(\tau)\cdot\mathrm{Log}\left(\mathbf{R}_{i+j-1}^{-1}\mathbf{R}_{i+j}\right)\right)
\end{equation}

\begin{equation}
    \label{Eqn:11}
    \dot{\mathbf{A}}_{j}=\mathbf{A}_{j}\dot{\tilde{\mathbf{B}}}(\tau)_{j}\mathrm{Log}\left(\mathbf{R}_{i+j-1}^{-1}\mathbf{R}_{i+j}\right)
\end{equation}

\begin{equation}
    \label{Eqn:12}
    \dot{\tilde{\mathbf{B}_{j}}}(\tau)=\dfrac{1}{\Delta t}\tilde{\mathbf{M}}^{(\text{3})}[0\enspace1\enspace2u\enspace3u^2]^\mathsf{T}%\begin{bmatrix}\text{0}\\[0.3em]\text{1}\\[0.3em]\text{2}u\\[0.3em]\text{3}u^\text{2}\end{bmatrix}
\end{equation}
\end{comment}

\begin{figure}[!t]
\centering
\includegraphics[width=.43\textwidth]{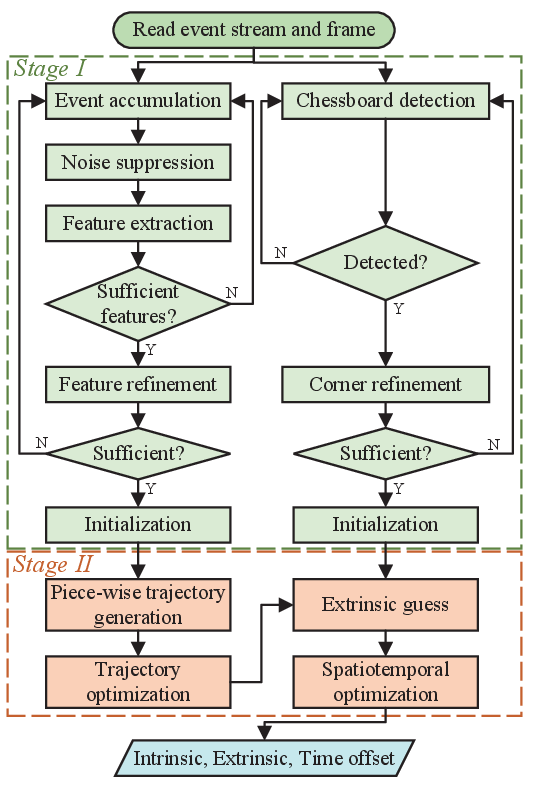}
\caption{Flowchart of the proposed calibration framework}
\label{Fig:framework}
\end{figure}

\section{Methodology}
\subsection{Calibration Framework}
For camera calibration, it is of utmost importance to accurately and robustly identify features on the calibration pattern. However, the checkerboard pattern \cite{Bib:Wang2022TIM}, which is widely used, is difficult to apply to event cameras because events disappear during parallel edge motion \cite{Bib:Gallego2020TPAMI}. Consequently, calibration of event cameras typically uses circular features. However, for moving frame-based cameras, circular features are more severely affected by motion blur than checkerboard crosspoints. To balance the needs of both event cameras and frame-based cameras, we designed a new calibration pattern that combines isotropic circles with checkerboard crosspoints, as illustrated in Fig. \ref{Fig:real-world}(a). The center of each circle in this pattern coincides with the center of the inner crosspoint. This hybrid pattern significantly enhances the recognition efficiency and accuracy of the event camera while maintaining compatibility with frame-based cameras.

Fig.~\ref{Fig:framework} presents the flowchart of the proposed calibration framework. In this letter, we divide the entire calibration process into two stages. The first stage focuses on feature extraction and refinement of the calibration pattern. The second stage is dedicated to optimizing the camera trajectory using piecewise B-splines to achieve accurate calibration results. These two stages will be elaborated in the following subsections.

\subsection{Event-Based Feature Recognizer}
\begin{figure*}[!t]
\centering
\includegraphics[width=0.9\textwidth]{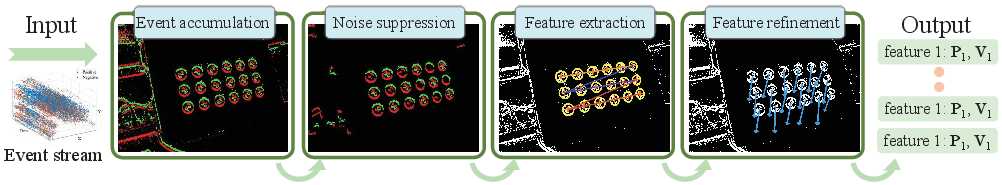}
\caption{The pipeline of event-based feature recognizer.}
\label{Fig:recognizer}
\end{figure*}

Unlike frame-based cameras, event cameras only output asynchronous events when the illumination of each pixel changes over a threshold, posing a challenge for robust calibration pattern recognition. To address this, we propose an event-based calibration pattern feature recognizer, illustrated in Fig.~\ref{Fig:recognizer}. First, we accumulate events over a short period of time $\Delta t$ according to their polarity to obtain ``\textit{accumulation frames}'' that resemble traditional images. The introduction of this ``\textit{accumulation frames}'' can help us to recognize the feature plate using some classical image processing algorithms. The following subsections describe the recognition algorithm based on ``\textit{accumulation frames}'' in detail.

\subsubsection{Noise Suppression}
Event cameras typically generate a large number of events from a static background when in motion. These noisy events do not contribute to the recognition of the calibration board and instead significantly increase the computational load on the feature extraction module, thereby reducing the efficiency of the calibration process. To address this, after obtaining the ``\textit{accumulation frames}'', a noise suppression module should be implemented to filter out most of the events that are unrelated to the calibration board.

For circular features, the triggered events typically consist of two semicircular arcs connecting regions of opposite polarity. However, many structures in the background have straight edges, making them more likely to have connected regions that resemble straight lines. To leverage this property, we introduce a fast and accurate connected component labeling (CCL) algorithm called BBDT \cite{Bib:Grana2010TIP}. This algorithm merges neighboring events with the same polarity to obtain all the connectivity regions. Then, the magnitudes of the two principal components of each connected region are calculated using PCA. For background-triggered connected regions, the magnitude of the second principal component $\Vert \textbf{PC}_\text{2} \Vert$ should be much smaller than the magnitude of the first principal component $\Vert \textbf{PC}_\text{1} \Vert$, so that a large number of noisy regions can be suppressed by the principal component magnitude ratio $\beta_{PC}$, as given that \mbox{$\beta_{PC}=\Vert \textbf{PC}_\text{1} \Vert / \Vert \textbf{PC}_\text{2} \Vert < T_{PC}$}, where $T_{PC}$ is a threshold for the $\beta_{PC}$, and any region with $\beta_{PC}$ higher than $T_{PC}$ is suppressed and not involved in subsequent operations.

\subsubsection{Feature Extraction}
Following noise suppression, we proceed to extract potential circular features from the remaining regions. Specifically, we identify two candidate regions of opposite polarity based on their distance. Subsequently, we fit the elliptic equation using all pixels contained within these regions, exploiting the fact that circular features adhere to the elliptic model under a projective transform. Then the fitting error $e_{fit}$ is calculated, excluding candidate regions with a fitting error exceeding the fitting threshold $T_{fit}$.

Furthermore, additional geometric constraints are needed to eliminate the remaining false positive candidate regions. Specifically, the PCA magnitudes of two connected components with opposite polarity, which together form the same ellipse, are calculated separately. For a true ellipse feature, these two semicircular arcs should exhibit similar PCA magnitudes; therefore, candidates with significant differences in PCA magnitudes are considered false positives. Additionally, the contribution of both connected components to the ellipse’s circumference should be nearly equal. This means that the angular range $\theta_r$ of the two candidate regions, relative to the center of the ellipse, should be close to 180$^{\circ}$.

Regions that successfully meet these geometric constraints are considered to accurately represent the elliptical features in the calibration pattern. The next step involves decoding the relative positions of the elliptical features using their spatial distribution to align with the frame-based camera’s recognized results. The calibration pattern’s default orientation is set so that the upper-left feature has black regions in the upper-left and lower-right directions. By identifying the elliptical feature in the upper-left corner of the current ``\textit{accumulation frames}'', the pattern’s orientation can be determined from the event distribution.

\begin{figure}[!t]
\centering
\includegraphics[width=.38\textwidth]{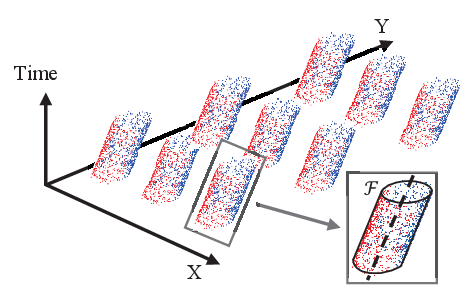}
\caption{Schematic of the moving ellipse model. The set of events belonging to the same elliptical feature can be considered as a three-dimensional oblique elliptical cylinder.}
\label{Fig:ellipse}
\end{figure}

\subsubsection{Feature Refinement}
Neglecting the timestamps of events during elliptical fitting inevitably introduces errors, thereby affecting the precision of the calibration results. However, by regarding the timestamps $t$ as a third dimension alongside the pixel coordinates $[x, y]^T$, events can be conceptualized as points within a three-dimensional space. Each ``\textit{accumulation frames}'', corresponding to a brief time interval, allows for the assumption of solely translational movement with speed $[v_x, v_y]^T$, parallel to the pixel plane, for each elliptical feature at any given moment within this interval, as demonstrated in Fig.~\ref{Fig:ellipse}. Consequently, the moving ellipse model $\mathcal{F}$ is described by the following representation:
\begin{equation}
    \label{Eqn:13}
    \mathcal{F}:\alpha x(t)^2+\beta x(t)y(t)+\gamma y(t)^2+\eta x(t)+\epsilon y(t)+\zeta=\text{0}
\end{equation}

In matrix form,
\begin{equation}
    \label{Eqn:14}
    \begin{cases}
        \mathcal{F}:\mathbf{P}(t)^{T}\mathbf{Q}\mathbf{P}(t)=\text{0} \\
        \mathbf{P}(t)=\mathbf{P}(t_{0})-\mathbf{V}\cdot(t-t_0)\\
        \mathbf{Q}=\begin{pmatrix}\alpha&\beta/\text{2}&\eta/\text{2}\\\beta/\text{2}&\gamma&\epsilon/\text{2}\\\eta/\text{2}&\epsilon/\text{2}&\zeta\end{pmatrix}
    \end{cases}
\end{equation}
where $\mathbf{P}(t)=[x(t), y(t), \text{1}]^T$, $\mathbf{V}=[v_x, v_y]^T$, $t \in [t_0, t_0 + \text{2}\delta_t]$, and $t_0$ represents the starting time of current ``\textit{accumulation frame}''. Notice that the moving ellipse model is optimized separately for each elliptic feature, and the cost function for model optimization is defined as
\begin{equation}
    \label{Eqn:15}
    \arg\min_{\mathcal{F}, \mathbf{V}}\sum_{i \in \{e\}}\left(\Vert\mathbf{P}_i^T\mathbf{Q}\mathbf{P}_i\Vert^1\right)
\end{equation}

An initial moving ellipse model is created by first calculating the average timestamps of the events and fitting the elliptical model using only their spatial coordinates. Then, by substituting the events into this model, we refine the elliptical features through optimization using the Levenberg-Marquardt algorithm.

\subsection{Trajectory Optimization}
\begin{figure}[!t]
\centering
\includegraphics[width=.4\textwidth]{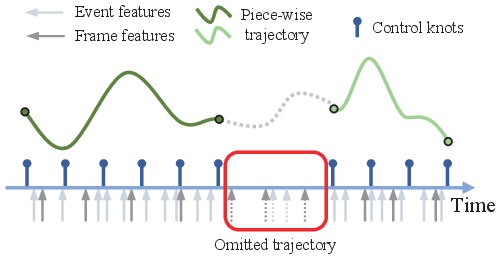}
\caption{Schematic of a piece-wise B-spline trajectory. The number of event features inside the red box is insufficient; therefore, this segment of the trajectory is omitted.}
\label{Fig:trajectory}
\end{figure}

After obtaining the refined features, the calibration parameters are optimized to yield the final results. Given the additional degrees of freedom introduced by the continuous-time trajectory, simultaneously optimizing all parameters risks the system becoming trapped in local minima. To address this, a two-stage optimization strategy is employed. Initially, the intrinsic parameters and motion trajectory of the event camera are optimized. These optimized parameters then serve as prior information to refine the remaining variables. The validity and superiority of this approach are demonstrated through the ablation study in Sec.~V.

The feature refinement process turns discrete elliptical features into densely populated patterns within the relevant time period, facilitating the optimization of event camera poses in a continuous-time trajectory. However, continuous visibility of the calibration pattern during the process is difficult to maintain, often resulting in recognition failures due to incomplete patterns in the event stream. To counter this, the trajectory is segmented based on recognizer output, and a piecewise B-spline-based optimizer for event camera pose trajectories is introduced.

First, based on the predefined knot interval $\Delta t$ and the results of the recognizer, the features whose timestamps differ from the timestamps of other features by more than $\Delta t$ are eliminated. In addition, segments containing too few features are also excluded to ensure optimization accuracy, and only the more desirable segments are preserved. Fig.~\ref{Fig:trajectory} illustrates the segmentation process of the trajectory. The entire calibration process $\Lambda_{\mathcal{L}}$ is divided into a combination of $M$ segments of trajectories $\mathcal{L}$:
\begin{equation}
    \label{Eqn:16}
    \Lambda_\mathcal{L}=\sum_m\{\mathcal{L}_m\}_{a_m}^{b_m},m=\text{1},\ldots,M
\end{equation}
where $a_m$ and $b_m$ are the starting and ending times corresponding to the $m$th segment of B-splines, respectively. Each segment of B-splines is optimized by only the features whose timestamps belong to its time period.

\begin{figure}[!t]
\centering
\includegraphics[width=.4\textwidth]{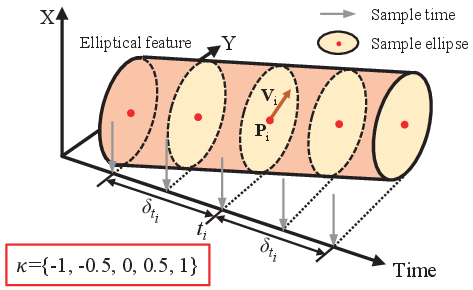}
\caption{Schematic of sampling on a continuously moving ellipse model.}
\label{Fig:refine}
\end{figure}

For $\mathcal{L}_m$, the corresponding state vector $\mathcal{X}_e$ is:
\begin{equation}
    \label{Eqn:17}
    \mathcal{X}_e=[\xi^\text{1}_m, \ \xi^\text{2}_m, \ \cdots, \ \xi^{N_m}_m, \ K_e, \ D_e]
\end{equation}
where $\xi_m^i$ is the control point of $\mathcal{L}_m$, $N_m$ is the number of control points, and $K_e$ and $D_e$ are the intrinsic and distortion parameters of the event camera. The corresponding visual residual for the $j$th feature based on reprojection error is defined as:

\begin{equation}
    \label{Eqn:18}
    \begin{aligned}
        \mathbf{r}_{e}(\mathcal{X}_e)=\sum_{k\in\mathcal{K}}\pi_{e}(\mathbf{R}_{w}^{e}(t_{i}+k\cdot\delta_{t_{i}})P_{j}^{w}+\mathbf{t}_{w}^{e}(t_{i}+k\cdot\delta_{t_{i}}))\\-(\begin{bmatrix}u_j^e\\v_j^e\end{bmatrix}+k\cdot\delta_{t_{i}}\mathbf{V}_i)
    \end{aligned}
\end{equation}
where $\mathbf{R}^e_w(\cdot)$ and $\mathbf{t}^e_w(\cdot)$ are derived from the B-spline trajectory using (\ref{Eqn:6}) and (\ref{Eqn:7}), respectively, and $P_{j}^{w}$ is the 3D spatial point of the $j$th feature in the world coordinate system, obtained from the geometric relationship of the calibration pattern. Since the feature refinement yields a continuous moving ellipse model, the residuals can be constructed by sampling the model at any time. Here, $\mathcal{K}$ denotes the partition of $\delta_{t_{i}}$, which defines the sampling interval of the feature, as shown in Fig.~\ref{Fig:refine}. The function $\pi_e(\cdot)$ projects the spatial point $P_{j}^{w}$ onto the ``\textit{accumulation frame}''.

The intrinsic and distortion parameters of the event camera, along with the control points of the splines, are jointly optimized by minimizing the following cost function:
\begin{equation}
    \label{Eqn:19}
    \arg\min_{\mathcal{X}_e}\left\{\sum\rho(\|\mathbf{r}_e(\mathcal{X}_e)\|^2)\right\}
\end{equation}
where $\rho(\cdot)$ is the Huber loss function, employed to reduce the impact of outliers. In the optimization problem, only the residuals corresponding to features whose timestamps lie within these piecewise trajectories are incorporated.

\subsection{Spatiotemporal Calibration}
The final step involves optimizing the intrinsic and extrinsic parameters of the frame-based camera, as well as the time offset between the two cameras, using the previously optimized trajectories. The corresponding state vector $\mathcal{X}_f$ is:

\begin{equation}
    \label{Eqn:20}
    \mathcal{X}_f=[K_f, \ D_f, \ \mathbf{T}^f_e, \ t_d]
\end{equation}
where $\mathbf{T}^f_e$ is the transformation matrix between the two cameras and $t_d$ is the difference between the real timestamps of the two cameras, i.e., the time offset.

Similarly, define the visual residuals based on reprojection errors in spatiotemporal calibration as:
\begin{equation}
    \label{Eqn:21}
    \mathbf{r}_f(\mathcal{X}_f)=\pi_f\left(\mathbf{R}^f_e\left(\mathbf{R}^e_w(t_i+t_d)P^w_j+\mathbf{t}^e_w(t_i+t_d)\right)+\mathbf{t}^f_e\right)-\begin{bmatrix}u^f_j\\[6pt] v^f_j\end{bmatrix}
\end{equation}
where $\mathbf{R}^e_w(\cdot) \in\mathbb{SO}$3 and $\mathbf{t}^e_w(\cdot) \in \mathbb{SO}$3 refer to the rotation and translation of the event camera pose represented by the B-spline, and $\pi_f(\cdot)$ projects the spatial points onto the image plane of the frame-based camera.

From (\ref{Eqn:21}), the Jacobian $J_{t_d}$ of $\mathbf{r}_f$ w.r.t $t_d$ can be obtained by the chain rule:
\begin{equation}
    \label{Eqn:22}
    J_{t_d} = \frac{\partial \mathbf{r}_f}{\partial P_j^f} \left( \frac{\partial P_j^f}{\partial t_d} \right) = \frac{\partial \mathbf{r}_f}{\partial P_j^e} \left( \frac{\partial P_j^f}{\partial \mathbf{R}_w^e} \frac{\partial \mathbf{R}_w^e}{\partial t_d} + \frac{\partial P_j^f}{\partial \mathbf{t}_w^e} \frac{\partial \mathbf{t}_w^e}{\partial t_d} \right)
\end{equation}
where $P_j^f=\mathbf{R}^f_e\left(\mathbf{R}^e_w(t_i+t_d)P^w_i+\mathbf{t}^e_w(t_i+t_d)\right)+\mathbf{t}^f_e$.

Based on (\ref{Eqn:8}), (\ref{Eqn:9}), and (\ref{Eqn:22}), the structure of $\partial P_i^f / \partial t_d$ can be derived straightforwardly:
\begin{equation}
    \label{Eqn:23}
    \frac{\partial P_j^f}{\partial \mathbf{R}_w^e}\frac{\partial \mathbf{R}_w^e}{\partial t_d}=\mathbf{R}^f_e \dot{\mathbf{R}}^e_w(t_i+t_d)P^w_j + \mathbf{R}^f_e \mathbf{v}^e_w(t_i+t_d)
\end{equation}

To jointly optimize the intrinsic and extrinsic parameters of the frame-based camera, as well as the time offset, the following cost function is minimized:
\begin{equation}
    \label{Eqn:24}
    \arg\min_{\mathcal{X}_f}\left\{\sum\rho(\|\mathbf{r}_f(\mathcal{X}_f)\|^2)\right\}
\end{equation}

In the optimization process, an initial value must be provided. Here, the initial value for the time offset $t_d$ is set to 0. We then iterate through the timestamps of each frame from the frame-based camera and each ``\textit{accumulation frame}'' from the event camera, identifying the pair with the closest timestamps. This pair is used to calculate for the corresponding extrinsic parameters, which are then used as the initial $\mathbf{T}^f_e$. Notably, the optimization problems in (\ref{Eqn:15}), (\ref{Eqn:19}), and (\ref{Eqn:24}) are solved using Google Ceres \cite{Bib:Ceres}.

\section{Experiments}
In this section, several experiments are conducted to evaluate the performance of EF-Calib, encompassing intrinsic calibration test, extrinsic calibration test, and time offset calibration test. Additionally, two ablation studies are conducted to evaluate the contribution of several key modules and the optimization strategy within \mbox{EF-Calib}.

\subsection{System Setup}
A real-world stereo vision system was designed as Fig.~\ref{Fig:real-world}(b) shows. It contains an event camera and a frame-based camera. The two cameras are integrated by a slide, on which the baseline and viewing angle can be arbitrarily adjusted to test the calibration performance of EF-Calib in different situations comprehensively. The event camera utilized in this letter is the Inivation DAVIS 346, featuring a resolution of 346$\times$260 and a maximum temporal resolution of 1~\textmu s. Additionally, this type of camera can also generate regular frames at a frequency of 30~Hz under standard illumination conditions. This configuration can be readily employed to compare \mbox{EF-Calib} with a high-quality, frame-based calibration pipeline, such as the OpenCV calibration toolbox \cite{Bib:Bradski2008}. The frame-based camera employed is the HikVision MV-CE013-80UM industrial camera with a global shutter and a resolution of 1280$\times$1024 pixels. Note that no hardware synchronization was utilized in the stereo vision system. This deliberate choice was made to provide a more rigorous evaluation of the calibration capability of EF-Calib under real-world conditions.

\begin{figure}[!t]
\centering
\includegraphics[width=.45\textwidth]{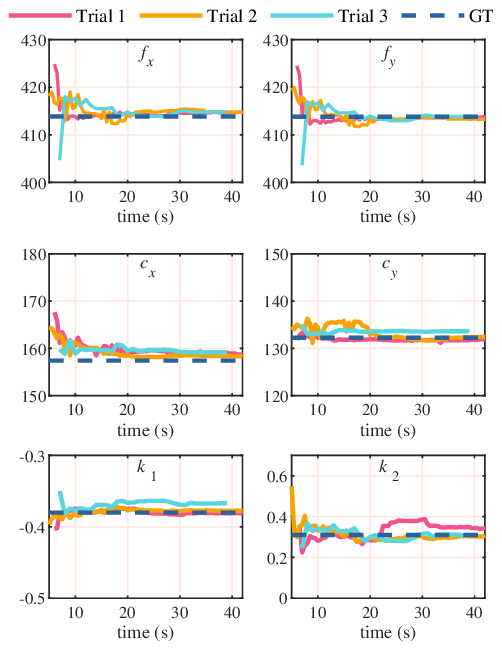}
\caption{Results of the intrinsic calibration test.}
\label{Fig:intrinsic_exp}
\end{figure}

\begin{table}[]
\renewcommand\arraystretch{1}
\setlength{\tabcolsep}{1mm}{}
\centering
\caption{Comparative Results of Intrinsic Calibration Test}
\label{Tab:Exp_1}
\begin{tabular}{c|c|c|c|c|c|c|c}
\hline
\hline
Methods              & $f_x$ & $f_y$ & $c_x$ & $c_y$ & $k_\text{1}$ & $k_\text{2}$ & RPE\\ \hline
Frame-based (GT)         & \textbf{413.84} & \textbf{413.80} & \textbf{157.42} & \textbf{132.25} & \textbf{-0.38} & \textbf{0.31} & \textbf{0.13} \\
E2Calib \cite{Bib:Muglikar2021CVPR}      & 417.56 & 417.24  & 159.86  & 132.35  & -0.36 & 0.09 & 0.41 \\
E-Calib \cite{Bib:Salah2023Arxiv}         & 404.66 & 403.99 & 159.81 & 132.69 & -0.37 & 0.32 & 0.33 \\
EF-Calib (Trial 1)   & \textbf{414.06} & \textbf{413.28} & \textbf{158.03} & \textbf{132.43} & \textbf{-0.38} & \textbf{0.31} & \textbf{0.10} \\
EF-Calib (Trial 2)   & \textbf{414.79} & \textbf{413.85} & \textbf{158.74} & \textbf{131.90} & \textbf{-0.38} & \textbf{0.34} & \textbf{0.12} \\
EF-Calib (Trial 3)   & \textbf{414.87} & \textbf{413.94} & \textbf{159.16} & \textbf{133.65} & \textbf{-0.37} & \textbf{0.31} & \textbf{0.13} \\\hline
\hline
\end{tabular}
\end{table}

\subsection{Calibration Experiments}
The calibration performance of a stereo vision system is usually affected by the camera baseline and viewing angle. To fully evaluate the calibration performance of EF-Calib, we conducted calibration experiments in three settings and analyzed the corresponding calibration results separately. In the first setting (Trial~1), the cameras are configured for a regular baseline. In the second setting (Trial~2), the cameras are configured for a wide baseline. In the third setup (Trial~3), the cameras are configured as a narrow baseline. For each trial, the cameras are adjusted to obtain a reasonable viewing angle, ensuring sufficient overlap of the camera FOV. For each trial, images with event data are recorded simultaneously for sufficient time ($\sim$40~s) to achieve converged calibration results.

\subsubsection{Intrinsic Calibration Test}
We utilized the event stream data from each of the three trials mentioned above to independently complete the intrinsic calibration of the event camera. Previously, we completed the intrinsic calibration using OpenCV toolbox \cite{Bib:Bradski2008} with the frame provided by DAVIS 346 and considered this calibration result as the ground truth. We obtained the intrinsic parameters and reprojection errors (RPE) of the event camera using EF-Calib and two state-of-the-art intrinsic calibration methods \cite{Bib:Muglikar2021CVPR, Bib:Salah2023Arxiv}, and compared them with the ground truth. Note that the calibration patterns used by the compared methods are the ones originally employed by them: \cite{Bib:Muglikar2021CVPR} utilizes a checkerboard pattern, while \cite{Bib:Salah2023Arxiv} employs an asymmetric circular pattern.

Table~\ref{Tab:Exp_1} shows the intrinsic calibration results for each method. As shown, the intrinsic parameters derived from our method are the closest to the ground truth (GT), with consistently low RPE across all trials, and the results are highly stable across the three trials. In addition, Fig.~\ref{Fig:intrinsic_exp} illustrates the plot of the intrinsic parameters over time. It can be noticed that EF-Calib can get converged results in less than 20~s, demonstrating the ease of use of our method.

\subsubsection{Extrinsic Calibration Test}
To evaluate the extrinsic calibration performance of the EF-Calib, we calculated the errors corresponding to rotation and translation separately for each trial, i.e.,
\begin{equation}
\label{Eqn:25}
    \begin{aligned}
        e_{t}&=\frac{1}{N}\sum_i^N\left\|\mathbf{t}^{e}_w(t_i+t_d)-\mathbf{T}_f^e{\mathbf{t}^f_{w}}_i\right\|_2  \\
        e_{r}&=\frac{1}{N}\sum_i^N\|\boldsymbol{\theta}(\mathbf{R}_w^e(t_i+t_d))-\boldsymbol{\theta}(\mathbf{R}_f^e{\mathbf{R}_w^f}_i)\|_2
    \end{aligned}
\end{equation}
where $N$ is the frame number and $\boldsymbol{\theta}(\cdot)$ represents the absolute angle corresponding to the rotation matrix. Similar to the intrinsic calibration test, we also acquired 30 pairs of images (each is captured statically, comprising one image from the frame-based camera and one image from the DAVIS~346) featuring a checkerboard calibration board, captured simultaneously from different viewing angles by the two cameras. These image pairs were then calibrated using the OpenCV toolbox to obtain the corresponding extrinsic parameters for both cameras. We refer to this method as ``\textit{frame-based}''. According to (\ref{Eqn:25}), we can calculate the rotation and translation errors corresponding to the “frame-based” method and use them as a reference to evaluate the extrinsic calibration accuracy of EF-Calib. From Table~\ref{Tab:Exp_2}, it can be seen that EF-Calib can achieve the same level of error as the frame-based extrinsic calibration, verifying its effectiveness in extrinsic calibration.

\begin{table}[!t]
\renewcommand\arraystretch{1}
\setlength{\tabcolsep}{2mm}{}
\centering
\caption{Comparative Results of Extrinsic Calibration Test}
\label{Tab:Exp_2}
\begin{tabular}{c|c|c|c|c}
\hline
\hline
Trial                    & Method      & $e_t$ (mm) & $e_r$ ($^{\circ}$) & Frames \\ \hline
\multirow{2}{*}{Trial 1} & Frame-based & 0.350 & 0.092 & 30  \\
                         & EF-Calib    & 0.534 & 0.198 & 250 \\ \hline
\multirow{2}{*}{Trial 2} & Frame-based & 0.420 & 0.185 & 30  \\
                         & EF-Calib    & 0.657 & 0.313 & 207 \\ \hline
\multirow{2}{*}{Trial 3} & Frame-based & 0.254 & 0.095 & 30  \\
                         & EF-Calib    & 0.364& 0.291 & 184  \\ \hline
\hline
\end{tabular}
\end{table}

\begin{figure}[!t]
\centering
\includegraphics[width=.46\textwidth]{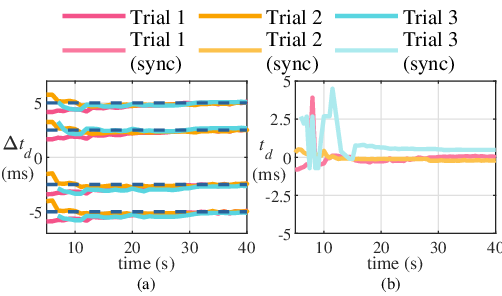}
\caption{Results of Time Offset Calibration Test. (a) Without hardware synchronization. (b) With hardware synchronization.}
\label{Fig:timeoffset_exp}
\end{figure}

\subsubsection{Time Offset Calibration Test}
Time offset estimation is crucial for multi-camera systems without hardware triggers. In this test, we evaluated EF-Calib's ability to calibrate the time offset. First, we compared the results with the original calibration by manually adjusting the timestamps. Specifically, the timestamps were modified by $\pm$2.5~ms and $\pm$5~ms for each trial, and the differences between the adjusted and original time delays were calculated and compared to the $\Delta t_d$. As shown in Fig.~\ref{Fig:timeoffset_exp}(a), EF-Calib accurately calibrates the time offset across different trials. Additionally, we synchronized the cameras' timestamps via hardware and then calibrated the time offset under the same settings to see if $t_d$ approaches 0~ms. Fig.~\ref{Fig:timeoffset_exp}(b) shows that EF-Calib successfully calibrates the time offset to near 0~ms after hardware synchronization.

\subsection{Ablation Studies}
\begin{table}[!t]
\renewcommand\arraystretch{1}
\setlength{\tabcolsep}{3pt}
\centering
\caption{Ablation Study of the Key Modules in EF-Calib}
\label{Tab:Exp_3}
\begin{tabular}{c|c|c|cc|cc|cc}
\hline
\hline
\multirow{2}{*}{\begin{tabular}[c]{@{}c@{}}Piece-wise\\trajectory\end{tabular}} & \multirow{2}{*}{\begin{tabular}[c]{@{}c@{}}Temporal\\calib\end{tabular}} & \multirow{2}{*}{\begin{tabular}[c]{@{}c@{}}Feature\\refine\end{tabular}} & \multicolumn{2}{c|}{Trial 1}     & \multicolumn{2}{c|}{Trial 2}     & \multicolumn{2}{c}{Trial 3}  \\ \cline{4-9}
 & & & \multicolumn{1}{c|}{$e_t$} & $e_r$ & \multicolumn{1}{c|}{$e_t$} & $e_r$ & \multicolumn{1}{c|}{$e_t$} & $e_r$ \\ \hline
 &  &  & \multicolumn{1}{c|}{8.163} & 2.863 & \multicolumn{1}{c|}{4.664} & 2.637 & \multicolumn{1}{c|}{1.216} & 1.451\\
\ding{52} & & & \multicolumn{1}{c|}{8.002} & 2.819 & \multicolumn{1}{c|}{3.879} & 2.786 & \multicolumn{1}{c|}{0.497} & 1.432\\
\ding{52} & \ding{52} & & \multicolumn{1}{c|}{0.914} & 5.691 & \multicolumn{1}{c|}{\textbf{0.632}} & 0.324 & \multicolumn{1}{c|}{0.371} & 0.307\\
\ding{52} & \ding{52} & \ding{52}  & \multicolumn{1}{c|}{\textbf{0.534}} & \textbf{0.198} & \multicolumn{1}{c|}{0.657} & \textbf{0.313} & \multicolumn{1}{c|}{\textbf{0.364}} & \textbf{0.291} \\ \hline
\hline
\end{tabular}
\end{table}

\begin{table}[!t]
\renewcommand\arraystretch{1}
\setlength{\tabcolsep}{4pt}
\centering
\caption{Ablation Study of Optimization Strategies}
\label{Tab:Exp_4}
\begin{tabular}{c|c|cc|cc|cc}
\hline
\hline
\multirow{2}{*}{\begin{tabular}[c]{@{}c@{}}Two-stage\end{tabular}} & \multirow{2}{*}{\begin{tabular}[c]{@{}c@{}}Separation\end{tabular}} & \multicolumn{2}{c|}{Trial 1}          & \multicolumn{2}{c|}{Trial 2}         & \multicolumn{2}{c}{Trial 3}          \\ \cline{3-8}
& & \multicolumn{1}{c|}{$e_t$} & $e_r$ & \multicolumn{1}{c|}{$e_t$} & $e_r$ & \multicolumn{1}{c|}{$e_t$} & $e_r$ \\ \hline
& & \multicolumn{1}{c|}{11.026} & 0.628 & \multicolumn{1}{c|}{1.429} & 0.351 & \multicolumn{1}{c|}{1.600} & 0.390 \\
\ding{52} & & \multicolumn{1}{c|}{0.552}  & \textbf{0.211} & \multicolumn{1}{c|}{0.697} & \textbf{0.320} & \multicolumn{1}{c|}{0.383} & \textbf{0.289} \\
\ding{52} & \ding{52} & \multicolumn{1}{c|}{\textbf{0.548}}  & 0.213 & \multicolumn{1}{c|}{\textbf{0.696}} & \textbf{0.320} & \multicolumn{1}{c|}{\textbf{0.382}} & \textbf{0.289} \\
\hline
\hline
\end{tabular}
\end{table}

To thoroughly analyze and validate the performance of each module within EF-Calib, as well as the optimization strategy employed, we conducted two ablation studies. The first study focused on three key modules: piece-wise trajectories, temporal calibration, and feature refinement. Table~\ref{Tab:Exp_3} shows the impact of these modules on the calibration error of EF-Calib’s extrinsic parameters. The Baseline algorithm, as listed in the top row, represents the entire event camera motion trajectory with a single continuous B-spline curve, assumes no time offset between cameras, and does not apply feature refinement. As Table~\ref{Tab:Exp_3} illustrates, incorporating these three modules significantly enhances calibration accuracy and ensures convergence to the correct solution.

The second ablation study compared different optimization strategies. Table~\ref{Tab:Exp_4} demonstrates that our strategy not only improves calibration accuracy over single-stage joint optimization but also significantly reduces computational load while maintaining accuracy comparable to the incremental two-stage optimization. This efficiency is attributed to the first stage’s effective refinement of parameters to near-optimal values, making the EF-Calib optimization strategy both highly accurate and efficient.

\section{Conclusion}
In this letter, we propose a novel calibration framework called EF-Calib, aiming to achieve spatiotemporal calibration of intrinsic parameters, extrinsic parameters, and time offset for event and frame-based cameras. Experimental results demonstrate that \mbox{EF-Calib} outperforms current state-of-the-art methods in intrinsic parameter estimation while also achieving high accuracy in extrinsic parameter and time offset estimation. These results demonstrate the spatiotemporal calibration capabilities of \mbox{EF-Calib} and lay a robust foundation for the fusion of event and frame.

In the future, we aim to explore markerless online calibration based on EF-Calib. Additionally, we plan to utilize \mbox{EF-Calib} to create novel visual perception frameworks that fuse events and frames.

\end{document}